\documentclass[11pt,a4paper]{article}
\usepackage[hyperref]{acl2018}
\usepackage{times}
\usepackage{latexsym}

\usepackage{url}
\usepackage{graphicx}
\usepackage{algorithm}
\usepackage{algorithmic}
\usepackage{booktabs}
\aclfinalcopy 

\title{Conditional BERT Contextual Augmentation}

\author{Xing Wu\textsuperscript{1,2}, \ Shangwen Lv\textsuperscript{1,2}, \ Liangjun Zang\textsuperscript{1}$^\dagger$, \ Jizhong Han\textsuperscript{1}, \ Songlin Hu\textsuperscript{1,2}$^\dagger$\\
	Institute of Information Engineering, Chinese Academy of Sciences, Beijing, China \\
  University of Chinese Academy of Sciences, Beijing, China \\
  {\tt \{wuxing,lvshangwen,zangliangjun,hanjizhong,husonglin\}@iie.ac.cn} \\}

\date{}
\pdfoutput=1
\begin{document}
\maketitle
\begin{abstract}
We propose a novel data augmentation method for labeled sentences called conditional BERT contextual augmentation. Data augmentation methods are often applied to prevent overfitting and improve generalization of deep neural network models. Recently proposed contextual augmentation augments labeled sentences by randomly replacing words with more varied substitutions predicted by language model. BERT demonstrates that a deep bidirectional language model is more powerful than either an unidirectional language model or the shallow concatenation of a forward and backward model. We retrofit BERT to conditional BERT by introducing a new conditional masked language model\footnote{The term ``conditional masked language model" appeared once in original BERT paper, which indicates context-conditional, is equivalent to term ``masked language model". In our paper, ``conditional masked language model" indicates we apply extra label-conditional constraint to the ``masked language model".} task. The well trained conditional BERT can be applied to enhance contextual augmentation. Experiments on six various different text classification tasks show that our method can be easily applied to both convolutional or recurrent neural networks classifier to obtain obvious improvement.
\end{abstract}

\section{Introduction}
Deep neural network-based models are easy to overfit and result in losing their generalization due to limited size of training data. In order to address the issue, data augmentation methods are often applied to generate more training samples. Recent years have witnessed great success in applying data augmentation in the field of speech area\cite{jaitly2013vocal,ko2015audio} and computer vision\cite{krizhevsky2012imagenet,simard1998transformation,szegedy2015going}. Data augmentation in these areas can be easily performed by transformations like resizing, mirroring, random cropping, and color shifting. However, applying these universal transformations to texts is largely randomized and uncontrollable, which makes it impossible to ensure the semantic invariance and label correctness. For example, given a movie review ``The actors is good", by mirroring we get ``doog si srotca ehT", or by random cropping we get ``actors is", both of which are meaningless. 

Existing data augmentation methods for text are often loss of generality, which are developed with handcrafted rules or pipelines for specific domains.
A general approach for text data augmentation is replacement-based method, which generates new sentences by replacing the words in the sentences with relevant words (e.g. synonyms). However, words with synonyms from a handcrafted lexical database likes WordNet\cite{miller1995wordnet} are very limited , and the replacement-based augmentation with synonyms can only produce limited diverse patterns from the original texts. To address the limitation of replacement-based methods, Kobayashi\cite{kobayashi2018contextual} proposed contextual augmentation for labeled sentences by offering a wide range of substitute words, which are predicted by a label-conditional bidirectional language model according to the context. But contextual augmentation suffers from two shortages: the bidirectional language model is simply shallow concatenation of a forward and backward model, and the usage of LSTM models restricts their prediction ability to a short range.

BERT, which stands for Bidirectional Encoder Representations from Transformers, pre-trained deep bidirectional representations by jointly conditioning on both left and right context in all layers. BERT addressed the unidirectional constraint by proposing a ``masked language model" (MLM) objective by masking some percentage of the input tokens at random, and predicting the masked words based on its context. This is very similar to how contextual augmentation predict the replacement words. But BERT was proposed to pre-train text representations, so MLM task is performed in an unsupervised way, taking no label variance into consideration. 

This paper focuses on the replacement-based methods, by proposing a novel data augmentation method called conditional BERT contextual augmentation. The method applies contextual augmentation by conditional BERT, which is fine-tuned on BERT. We adopt BERT as our pre-trained language model with two reasons. First, BERT is based on Transformer. Transformer provides us with a more structured memory for handling long-term dependencies in text. Second, BERT, as a deep bidirectional model, is strictly more powerful than the shallow concatenation of a left-to-right and right-to left model.
So we apply BERT to contextual augmentation for labeled sentences, by offering a wider range of substitute words predicted by the masked language model task. However, the masked language model predicts the masked word based only on its context, so the predicted word maybe incompatible with the annotated labels of the original sentences. In order to address this issue, we introduce a new fine-tuning objective: the "conditional masked language model"(C-MLM). The conditional masked language model randomly masks some of the tokens from an input, and the objective is to predict a label-compatible word based on both its context and sentence label. Unlike Kobayashi's work, the C-MLM objective allows a deep bidirectional representations by jointly conditioning on both left and right context in all layers. In order to evaluate how well our augmentation method improves performance of deep neural network models, following Kobayashi\cite{kobayashi2018contextual}, we experiment it on two most common neural network structures, LSTM-RNN and CNN, on text classification tasks. Through the experiments on six various different text classification tasks, we demonstrate that the proposed conditional BERT model augments sentence better than baselines, and conditional BERT contextual augmentation method can be easily applied to both convolutional or recurrent neural networks classifier. We further explore our conditional MLM task’s connection with style transfer task and demonstrate that our conditional BERT can also be applied to style transfer too.

Our contributions are concluded as follows:
\begin{itemize}
\item We propose a conditional BERT contextual augmentation method. The method allows BERT to augment sentences without breaking the label-compatibility. Our conditional BERT can further be applied to style transfer task.
\item Experimental results show that our approach obviously outperforms existing text data augmentation approaches.
\end{itemize}
To our best knowledge, this is the first attempt to alter BERT
to a conditional BERT or apply BERT on text generation tasks.

\section{Related Work}
\subsection{Fine-tuning on Pre-trained Language Model}
Language model pre-training has attracted wide attention and fine-tuning on pre-trained language model has shown to be effective for improving many downstream natural language processing tasks. Dai\cite{dai2015semi} pre-trained unlabeled data to improve Sequence Learning with recurrent networks. Howard\cite{howard2018universal} proposed a general transfer learning method, Universal Language Model Fine-tuning (ULMFiT), with the key techniques for fine-tuning a language model. Radford\cite{radford2018improving} proposed that by generative pre-training of a language model on a diverse corpus of unlabeled text, large gains on a diverse range of tasks could be realized. Radford\cite{radford2018improving} achieved large improvements on many sentence-level tasks from the GLUE benchmark\cite{wang2018glue}. BERT\cite{devlin2018bert} obtained new state-of-the-art results on a broad range of diverse tasks. BERT pre-trained deep bidirectional representations which jointly conditioned on both left and right context in all layers, following by discriminative fine-tuning on each specific task. Unlike previous works fine-tuning pre-trained language model to perform discriminative tasks, we aim to apply pre-trained BERT on generative tasks by perform the masked language model(MLM) task. To generate sentences that are compatible with given labels, we retrofit BERT to conditional BERT, by introducing a conditional masked language model task and fine-tuning BERT on the task. 

\subsection{Text Data Augmentation}
Text data augmentation has been extensively studied in natural language processing. Sample-based methods includes downsampling from the majority classes and oversampling from the minority class, both of which perform weakly in practice. Generation-based methods employ deep generative models such as GANs\cite{goodfellow2014generative} or VAEs\cite{bowman2015generating,hu2017toward}, trying to generate sentences from a continuous space with desired attributes of sentiment and tense. However, sentences generated in these methods are very hard to guarantee the quality both in label compatibility and sentence readability. In some specific areas \cite{jia2017adversarial,xie2017data,ebrahimi2017hotflip}.
word replacement augmentation was applied. Wang\cite{wang2015s} proposed the use of neighboring words in continuous representations to create new instances for every word in a tweet to augment the training dataset. Zhang\cite{zhang2015character} extracted all replaceable words from the given text and randomly choose $r$ of them to be replaced, then substituted the replaceable words with synonyms from WordNet\cite{miller1995wordnet}. Kolomiyets\cite{kolomiyets2011model} replaced only the headwords under a task-specific assumption that temporal trigger words usually occur as headwords. Kolomiyets\cite{kolomiyets2011model} selected substitute words with top-$K$ scores given by the Latent Words LM\cite{deschacht2009semi}, which is a LM based on fixed length contexts. Fadaee\cite{fadaee2017data} focused on the rare word problem in machine translation, replacing words in a source sentence with only rare words. A word in the translated sentence is also replaced using a word alignment method and a rightward LM. 
The work most similar to our research is Kobayashi\cite{kobayashi2018contextual}. Kobayashi used a fill-in-the-blank context for data augmentation by replacing every words in the sentence with language model. In order to prevent the generated words from reversing the information related to the labels of the sentences, Kobayashi\cite{kobayashi2018contextual} introduced a conditional constraint to control the replacement of words. Unlike previous works, we adopt a deep bidirectional language model to apply replacement, and the attention mechanism within our model allows a more structured memory for handling long-term dependencies in text, which resulting in more general and robust improvement on various downstream tasks.

\section{Conditional BERT Contextual Augmentation}
\subsection{Preliminary: Masked Language Model Task}
\subsubsection{Bidirectional Language Model} In general, the language model(LM) models the probability of generating natural language sentences or documents. Given a sequence $\textbf{\textit{S}}$ of N tokens, $<t_1,t_2,...,t_N>$, a forward language model allows us to predict the probability of the sequence as:
\begin{equation}
p(t_1,t_2,...,t_N) = \prod_{i=1}^{N}p(t_i|t_1, t_2,..., t_{i-1}).
\end{equation}
Similarly, a backward language model allows us to predict the probability of the sentence as:
\begin{equation}
p(t_1,t_2,...,t_N) = \prod_{i=1}^{N}p(t_i|t_{i+1}, t_{i+2},..., t_N).
\end{equation}
Traditionally, a bidirectional language model a shallow concatenation of independently trained forward and backward LMs.

\subsubsection{Masked Language Model Task}
In order to train a deep bidirectional language model, BERT proposed Masked Language Model (MLM) task, which was also referred to Cloze Task\cite{taylor1953Cloze}. MLM task randomly masks some percentage of the input tokens, and then predicts only those masked tokens according to their context.
Given a masked token ${t_i}$, the context is the tokens surrounding token ${t_i}$ in the sequence $\textbf{\textit{S}}$, i.e. cloze sentence ${\textbf{\textit{S}}\backslash \lbrace t_i \rbrace}$. The final hidden vectors corresponding to the mask tokens are fed into an output softmax over the vocabulary to produce words with a probability distribution ${p(\cdot|\textbf{\textit{S}}\backslash \lbrace t_i \rbrace)}$. MLM task only predicts the masked words rather than reconstructing the entire input, which suggests that more pre-training steps are required for the model to converge. Pre-trained BERT can augment sentences through MLM task, by predicting new words in masked positions according to their context.

\subsection{Conditional BERT}
\begin{figure*}[htp!]
	\begin{centering}
	\includegraphics[width=0.6\textwidth]{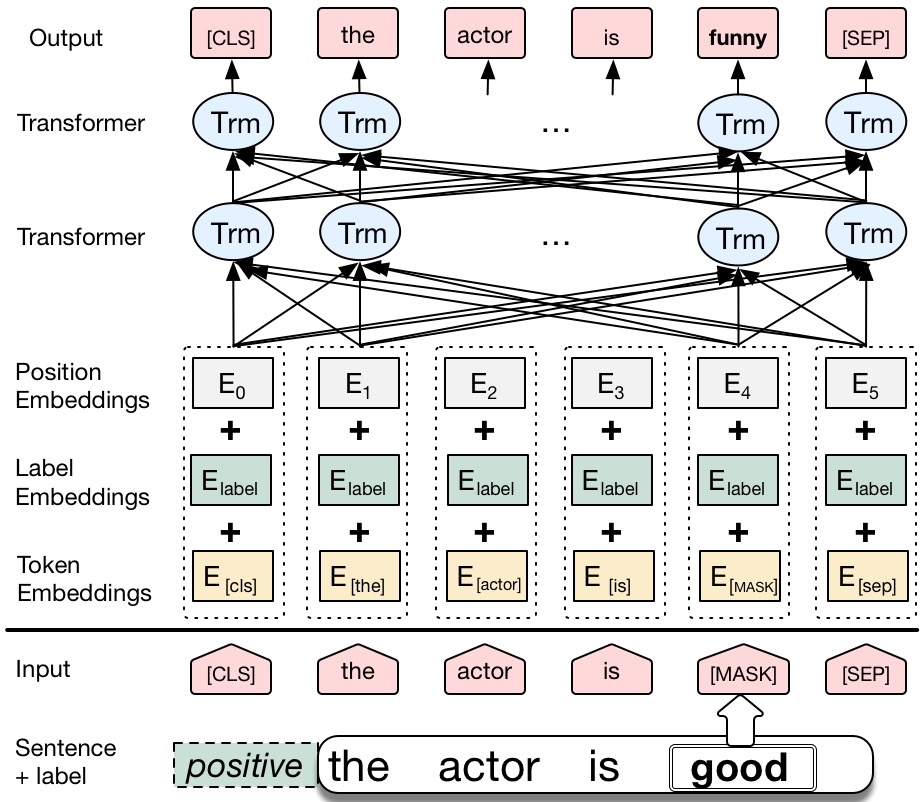}
	\caption{Model architecture of conditional BERT. The label embeddings in conditional BERT corresponding to segmentation embeddings in BERT, but their functions are different.} \label{fig1}
	\end{centering}
\end{figure*}

As shown in Fig \ref{fig1}, our conditional BERT shares the same model architecture with the original BERT. The differences are the input representation and training procedure. 

The input embeddings of BERT are the sum of the token embeddings, the segmentation embeddings and the position embeddings. For the segmentation embeddings in BERT, a learned sentence A embedding is added to every token of the first sentence, and if a second sentence exists, a sentence B embedding will be added to every token of the second sentence. However, the segmentation embeddings has no connection to the actual annotated labels of a sentence, like sense, sentiment or subjectivity, so predicted word is not always compatible with annotated labels. For example, given a positive movie remark ``this actor is good", we have the word ``good" masked. Through the Masked Language Model task by BERT, the predicted word in the masked position has potential to be negative word likes "bad" or "boring". Such new generated sentences by substituting masked words are implausible with respect to their original labels, which will be harmful if added to the corpus to apply augmentation. In order to address this issue, we propose a new task: ``conditional masked language model".
\subsubsection{Conditional Masked Language Model} The conditional masked language model randomly masks some of the tokens from the labeled sentence, and the objective is to predict the original vocabulary index of the masked word based on both its context and its label. Given a masked token ${t_i}$, the context ${\textbf{\textit{S}}\backslash \lbrace t_i \rbrace}$ and label ${y}$ are both considered, aiming to calculate ${p(\cdot|y,\textbf{\textit{S}}\backslash \lbrace t_i \rbrace)}$, instead of calculating ${p(\cdot|\textbf{\textit{S}}\backslash \lbrace t_i \rbrace)}$. Unlike MLM pre-training, the conditional MLM objective allows the representation to fuse the context information and the label information, which allows us to further train a label-conditional deep bidirectional representations.

To perform conditional MLM task, we fine-tune on pre-trained BERT. We alter the segmentation embeddings to label embeddings, which are learned corresponding to their annotated labels on labeled datasets. Note that the BERT are designed with segmentation embedding being embedding A or embedding B, so when a downstream task dataset with more than two labels, we have to adapt the size of embedding to label size compatible. We train conditional BERT using conditional MLM task on labeled dataset. After the model has converged, it is expected to be able to predict words in masked position both considering the context and the label.

\subsection{Conditional BERT Contextual Augmentation}
After the conditional BERT is well-trained, we utilize it to augment sentences. Given a labeled sentence from the corpus, we randomly mask a few words in the sentence. Through conditional BERT, various words compatibly with the label of  the sentence are predicted by conditional BERT. After substituting the masked words with predicted words, a new sentences is generated, which shares similar context and same label with original sentence. Then new sentences are added to original corpus. We elaborate the entire process in algorithm \ref{alg1}.

\begin{algorithm}
  \caption{Conditional BERT contextual augmentation algorithm. Fine-tuning on the pre-trained BERT , we retrofit BERT to conditional BERT using conditional MLM task on labeled dataset. After the model converged, we utilize it to augment sentences. New sentences are added into dataset to augment the dataset.}\label{alg1}
  \begin{algorithmic}[1]
  \STATE Alter the segmentation embeddings to label embeddings
  \STATE Fine-tune the pre-trained BERT using conditional MLM task on labeled dataset D until convergence
  \FOR {each iteration i=1,2,...,M}
  	\STATE Sample a sentence $s$ from D
  	\STATE Randomly mask $k$ words
  	\STATE Using fine-tuned conditional BERT to predict label-compatible words on masked positions to generate a new sentence $S^{\prime}$
  \ENDFOR
  \STATE Add new sentences into dataset $D$ to get augmented dataset $D^{\prime}$
  \STATE Perform downstream task on augmented dataset $D^{\prime}$
  \end{algorithmic}
\end{algorithm}

\section{Experiment}
In this section, we present conditional BERT parameter settings and, following Kobayashi\cite{kobayashi2018contextual}, we apply different augmentation methods on two types of neural models through six text classification tasks. The pre-trained BERT model we used in our experiment is BERT$_{BASE}$, with number of layers (i.e., Transformer blocks) $L = 12$, the hidden size $ H = 768$, and the number of self-attention heads $A = 12$, total parameters $= 110M$. Detailed pre-train parameters setting can be found in original paper\cite{devlin2018bert}. For each task, we perform the following steps independently. First, we evaluate the augmentation ability of original BERT model pre-trained on MLM task. We use pre-trained BERT to augment dataset, by predicted masked words only condition on context for each sentence. Second, we fine-tune the original BERT model to a conditional BERT. Well-trained conditional BERT augments each sentence in dataset by predicted masked words condition on both context and label. Third, we compare the performance of the two methods with Kobayashi's\cite{kobayashi2018contextual} contextual augmentation results. 
Note that the original BERT’s segmentation embeddings layer is compatible with two-label dataset. When the task-specific dataset is with more than two different labels, we should re-train a label size compatible label embeddings layer instead of directly fine-tuning the pre-trained one.

\subsection{Datasets}
Six benchmark classification datasets are listed in table \ref{tab1}. Following Kim\cite{kim2014convolutional}, for a dataset without validation data, we use 10\% of its training set for the validation set. Summary statistics of six classification datasets are shown in table 1.

\begin{table}[htp]
	\begin{centering}
		\caption{Summary statistics for the datasets after tokenization. $c$: Number of target classes. $l$: Average sentence length. $N$: Dataset size. $|V|$: Vocabulary size. $Test$: Test set size (CV means there was no standard train/test split and thus 10-fold cross-validation was used).}
		\setlength{\tabcolsep}{3mm}{\begin{tabular}{c||c|c|c|c|c}
		\hline
		\textbf{Data}& $c$& $l$& $N$& $|V|$& $Test$\\
		\hline
		SST5&	5&	18&		 11855&	17836& 2210\\
		SST2&	2&	19&		 9613&	16185& 1821\\
		Subj&	2&	23&		 10000&	21323& CV\\
		TREC&	6&	10&		 5952&	9592& 500\\
		MPQA&	2&	3&		 10606&	6246& CV\\
		RT&		2&	21&		 10662&	20287& CV\\
		\hline
		\end{tabular}}
		
		\label{tab1}
	\end{centering}
\end{table}

\noindent\textbf{SST}\cite{socher2013recursive} SST (Stanford Sentiment Treebank) is a dataset for sentiment classification on movie reviews, which are annotated with five labels (SST5: very positive, positive, neutral, negative, or very negative) or two labels (SST2: positive or negative).\\
\textbf{Subj}\cite{pang2004sentimental} Subj (Subjectivity dataset) is annotated with whether a sentence is subjective or objective.\\ 
\textbf{MPQA}\cite{wiebe2005annotating} MPQA Opinion Corpus is an opinion polarity detection dataset of short phrases rather than sentences, which contains news articles from a wide variety of news sources manually annotated for opinions and other private states (i.e., beliefs, emotions, sentiments, speculations, etc.).\\
\textbf{RT}\cite{pang2005seeing} RT is another movie review sentiment dataset contains a collection of short review excerpts from Rotten Tomatoes collected by Bo Pang and Lillian Lee. \\
\textbf{TREC}\cite{li2002learning} TREC is a dataset for classification of the six question types (whether the question is about person, location, numeric information, etc.). 

\subsection{Text classification}

\begin{table*}[htp]
	\begin{centering}
		\caption{Accuracies of different methods for various benchmarks on two classifier architectures. C-BERT, which represents conditional BERT, performs best on two classifier structures over six datasets. ``w/" represents ``with", lines marked with ``*" are experiments results from Kobayashi\cite{kobayashi2018contextual}.}
		\begin{tabular}{c|cccccc|c}
		\hline
		Model& SST5& SST2& Subj& MPQA& RT& TREC& Avg.\\
		\hline
		CNN*& 41.3& 79.5& 92.4& 86.1& 75.9& 90.0& 77.53\\
		w/synonym*& 40.7& 80.0& 92.4& 86.3& 76.0& 89.6& 77.50\\
		w/context*& 41.9& 80.9& 92.7& 86.7& 75.9& 90.0& 78.02\\
		w/context+label*& 42.1& 80.8& 93.0& 86.7& 76.1& 90.5& 78.20\\
		w/BERT& 41.5& 81.9& 92.9& 87.7& 78.2& 91.8& 79.00\\
		w/C-BERT& \textbf{42.3}& \textbf{82.1}& \textbf{93.4}& \textbf{88.2}& \textbf{79.0}& \textbf{92.6}& \textbf{79.60}\\
		\hline
		RNN*& 40.2& 80.3& 92.4& 86.0& 76.7& 89.0& 77.43\\
		w/synonym*& 40.5& 80.2& 92.8& 86.4& 76.6& 87.9& 77.40\\
		w/context*& 40.9& 79.3& 92.8& 86.4& 77.0& 89.3& 77.62\\
		w/context+label*& 41.1& 80.1& 92.8& 86.4& 77.4& 89.2& 77.83\\
		w/BERT& 41.3& 81.4& 93.5& 87.3& 78.3& 89.8& 78.60\\
		w/C-BERT& \textbf{42.6}& \textbf{81.9}& \textbf{93.9}& \textbf{88.0}& \textbf{78.9}& \textbf{91.0}& \textbf{79.38}\\
		\hline
		\end{tabular}
		
		\label{tab2}
	\end{centering}
\end{table*}

\subsubsection{Sentence Classifier Structure}
We evaluate the performance improvement brought by conditional BERT contextual augmentation on sentence classification tasks, so we need to prepare two common sentence classifiers beforehand. For comparison, following Kobayashi\cite{kobayashi2018contextual}, we adopt two typical classifier architectures: CNN or LSTM-RNN.  The CNN-based classifier\cite{kim2014convolutional} has convolutional filters of size {3, 4, 5} and word embeddings. All outputs of each filter are concatenated before applied with a max-pooling over time, then fed into a two-layer feed-forward network with ReLU, followed by the softmax function. An RNN-based classifier has a single layer LSTM and word embeddings, whose output is fed into an output affine layer with the softmax function. For both the architectures, dropout\cite{srivastava2014dropout} and Adam optimization\cite{kingma2014adam} are applied during training. The train process is finish by early stopping with validation at each epoch.

\subsubsection{Hyper-parameters Setting}
Sentence classifier hyper-parameters including learning rate, embedding dimension, unit or filter size, and dropout ratio, are selected using grid-search for each task-specific dataset. We refer to Kobayashi's implementation in the released code\footnote{\url{https://github.com/pfnet\-research/contextual\_augmentation}}.
For BERT, all hyper-parameters are kept the same as Devlin\cite{devlin2018bert}, codes in Tensorflow\footnote{\url{https://github.com/google-research/bert}} and PyTorch\footnote{\url{https://github.com/huggingface/pytorch-pretrained-BERT}} are all available on github and pre-trained BERT model can also be downloaded. The number of conditional BERT training epochs ranges in [1-50] and number of masked words ranges in [1-2].

\subsubsection{Baselines}
We compare the performance improvements obtained by our proposed method with the following baseline methods, ``w/" means ``with": 
\begin{itemize}
\item w/synonym: Words are randomly replaced  with synonyms from WordNet\cite{miller1995wordnet}. 
\item w/context: Proposed by Kobayashi\cite{kobayashi2018contextual}, which used a bidirectional language model to apply contextual augmentation, each word was replaced with a probability. 
\item w/context+label: Kobayashi’s contextual augmentation method\cite{kobayashi2018contextual} in a label-conditional LM architecture.
\end{itemize}

\subsubsection{Experiment Results}
Table \ref{tab2} lists the accuracies of the all methods on two classifier architectures. The results show that, for various datasets on different classifier architectures, our conditional BERT contextual augmentation improves the model performances most. BERT can also augments sentences to some extent, but not as much as conditional BERT does. For we masked words randomly, the masked words may be label-sensitive or label-insensitive. If label-insensitive words are masked, words predicted through BERT may not be compatible with original labels. The improvement over all benchmark datasets also shows that conditional BERT is a general augmentation method for multi-labels sentence classification tasks. 

\subsubsection{Effect of Number of Fine-tuning Steps}
We also explore the effect of number of training steps to the performance of conditional BERT data augmentation. The fine-tuning epoch setting ranges in [1-50], we list the fine-tuning epoch of conditional BERT to outperform BERT for various benchmarks in table \ref{tab3}. The results show that our conditional BERT contextual augmentation can achieve obvious performance improvement after only a few fine-tuning epochs, which is very convenient to apply to downstream tasks.
\begin{table}[htp]
	\begin{centering}
		\caption{Fine-tuning epochs of conditional BERT to outperform BERT for various benchmarks}
		\setlength{\tabcolsep}{1mm}{\begin{tabular}{c|cccccc}
		\hline
		Model& SST5& SST2& Subj& MPQA& RT& TREC\\
		\hline
		CNN& 4& 3& 1& 2& 2& 1\\
		\hline
		RNN& 6& 2& 2& 2& 1& 1\\
		\hline
		\end{tabular}}
		
		\label{tab3}
	\end{centering}
\end{table}

\section{Connection to Style Transfer}

\begin{table*}[htbp]
\begin{centering}
\caption{Examples generated by conditional BERT on the SST2 dataset. To perform style transfer, we reverse the original label of a sentence, and conditional BERT output a new label compatible sentence.}\label{tab4}
\setlength{\tabcolsep}{0.5mm}{
\begin{tabular}{l|l}
\toprule
\textbf{Original}:& there 's no disguising this as one of the worst films of the summer . \\
\textbf{Generated}:& there 's no disguising this as one of the best films of the summer . \\
\textbf{Original}:& it 's probably not easy to make such a worthless film ... \\
\textbf{Generated}:& it 's probably not easy to make such a stunning film ... \\
\textbf{Original}:& woody allen has really found his groove these days . \\
\textbf{Generated}:& woody allen has really lost his groove these days . \\
\bottomrule
\end{tabular}}
\end{centering}
\end{table*}

In this section, we further deep into the connection to style transfer and apply our well trained conditional BERT to style transfer task.
Style transfer is defined as the task of rephrasing the text to contain specific stylistic properties without changing the intent or affect within the context\cite{prabhumoye2018style}. Our conditional MLM task changes words in the text condition on given label without changing the context. View from this point, the two tasks are very close. So in order to apply conditional BERT to style transfer task, given a specific stylistic sentence, we break it into two steps: first, we find the words relevant to the style; second, we mask the style-relevant words, then use conditional BERT to predict new substitutes with sentence context and target style property. In order to find style-relevant words in a sentence, we refer to Xu\cite{xu2018unpaired}, which proposed an attention-based method to extract the contribution of each word to the sentence sentimental label.
For example, given a positive movie remark ``This movie is funny and interesting", we filter out the words contributes largely to the label and mask them. Then through our conditional BERT contextual augmentation method, we fill in the masked position by predicting words conditioning on opposite label and sentence context, resulting in ``This movie is boring and dull". The words ``boring" and ``dull" contribute to the new sentence being labeled as negative style. We sample some sentences from dataset SST2, transferring them to the opposite label, as listed in table \ref{tab4}.

\section{Conclusions and Future Work}
In this paper, we fine-tune BERT to conditional BERT by introducing a novel conditional MLM task. After being well trained, the conditional BERT can be applied to data augmentation for sentence classification tasks. Experiment results show that our model outperforms several baseline methods obviously. Furthermore, we demonstrate that our conditional BERT can also be applied to style transfer task. In the future, (1)We will explore how to perform text data augmentation on imbalanced datasets with pre-trained language model, (2) we believe the idea of conditional BERT contextual augmentation is universal and will be applied to paragraph or document level data augmentation.

\bibliography{Conditional_BERT_Contextual_Augmentation}
\bibliographystyle{acl_natbib}

\end{document}